\pdfoutput=1

\documentclass[11pt]{article}

\usepackage[final]{acl}

\usepackage{ulem}

\usepackage{times}
\usepackage{latexsym}
\usepackage{xspace}

\usepackage{amsmath}   
\usepackage{amssymb}   
\usepackage{amsfonts}  
\usepackage{titletoc}
\usepackage{tocloft}
\usepackage{multirow}

\usepackage{algorithm}
\usepackage{algorithmicx}
\usepackage{algpseudocode}

\usepackage[T1]{fontenc}

\usepackage[utf8]{inputenc}

\usepackage{microtype}

\usepackage{inconsolata}

\usepackage{graphicx}
\usepackage{booktabs}
\usepackage{xcolor}
\makeatletter
\renewcommand\@makefnmark{%
  \hbox{\textsuperscript{\normalfont\color{black}\@thefnmark}}%
}
\makeatother
\newcommand{\ours}{\textsc{Video-SkoT}\xspace}

\title{
\textsc{Video-Skill-CoT}: Skill-based Chain-of-Thoughts\\
for Domain-Adaptive Video Reasoning 
}

\author{%
Daeun Lee$^{1, }$\thanks{Equal contribution.} \quad
Jaehong Yoon$^{1,2, *}$ \quad
Jaemin Cho$^{1}$ \quad
Mohit Bansal$^{1}$\\
$^{1}$UNC Chapel Hill\quad\quad $^{2}$Nanyang Technological University\\
\texttt{\{daeun,jmincho,mbansal\}@cs.unc.edu}\quad \texttt{jaehong.yoon@ntu.edu.sg}\\
\url{https://video-skill-cot.github.io/}
}

\definecolor{darkgreen}{rgb}{0,0.5,0}
\definecolor{azureblue}{rgb}{0,0.5,1}
\definecolor{darkgreen}{rgb}{1,0,0}
\definecolor{color1}{HTML}{006EB8}
\definecolor{color2}{HTML}{009B55}
\definecolor{color3}{HTML}{924DBF}

\usepackage{pifont}

\newcommand{\xmark}{\ding{55}}

\hypersetup{
    colorlinks=true,
    linkcolor=magenta,
    filecolor=magenta,  
    urlcolor=magenta,
    citecolor=color3,
    pdftitle={Overleaf Example},
    pdfpagemode=FullScreen,
    }

\usepackage[capitalize]{cleveref}
\crefname{section}{Sec.}{Secs.}
\Crefname{section}{Section}{Sections}
\Crefname{table}{Table}{Tables}
\crefname{table}{Tab.}{Tabs.}
\crefname{appendix}{Sec.}{Secs.}
\Crefname{appendix}{Section}{Sections}

\usepackage{color,colortbl}
\definecolor{darkgreen}{rgb}{0,0.5,0}

\definecolor{darkgreen}{rgb}{1,0,0}

\definecolor{azureblue}{rgb}{0,0.5,1}

\definecolor{lightblue}{rgb}{0.90, 0.95, 0.98}
\newcolumntype{a}{>{\columncolor{lightblue}}c}

\begin{document}

\maketitle

\begin{abstract}
Recent advances in Chain-of-Thought (CoT) reasoning have improved complex video understanding, but existing methods often struggle to adapt to domain-specific skills (e.g., event detection, spatial relation understanding, emotion understanding) over various video content.
To address this, we propose
\textsc{Video-Skill-CoT} (a.k.a. \textsc{Video-SkoT})
a framework that automatically constructs and leverages skill-aware CoT supervisions for domain-adaptive video reasoning.
First, we construct skill-based CoT annotations: We extract domain-relevant reasoning skills from training questions, cluster them into a shared skill taxonomy, and create detailed multi-step CoT rationale tailored to each video-question pair for training.
Second, we introduce a skill-specific expert learning framework. Each expert module specializes in a subset of reasoning skills and is trained with lightweight adapters using the collected CoT supervision. 
We demonstrate the effectiveness of the proposed approach on three video understanding benchmarks, where
\ours consistently outperforms strong baselines. 
We also provide in-depth analyses on comparing different CoT annotation pipelines
and learned skills over multiple video domains.
\end{abstract} 


\begin{figure*}[t]
    \centering
    {
    \includegraphics[width=.99\textwidth]{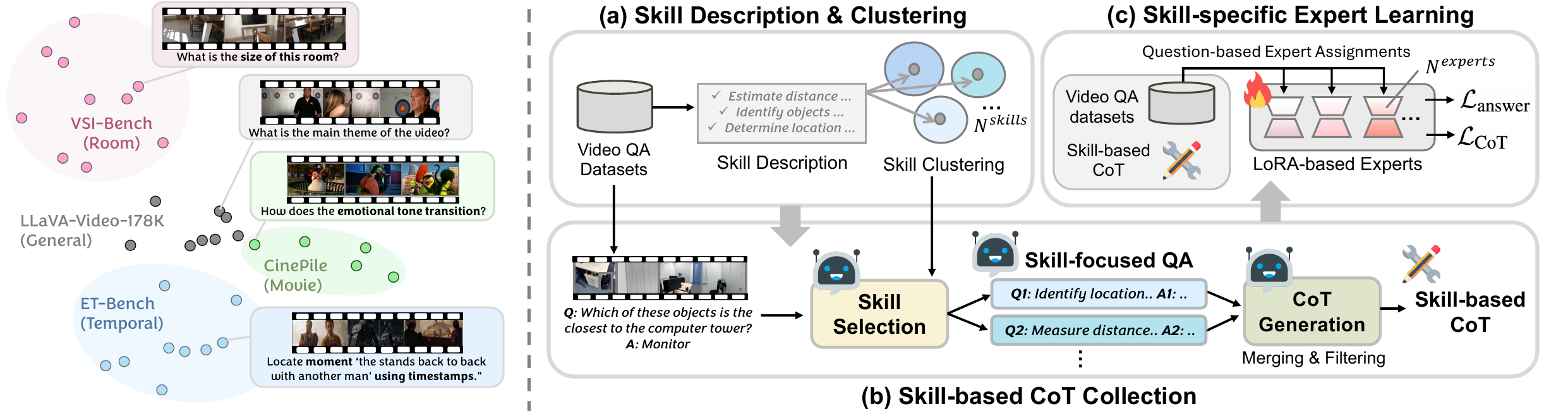}}
    \caption{
    \textbf{Left}: Video datasets require different reasoning skills.
    \textbf{Right}: \ours{} that automatically constructs and leverages skill-aware CoT supervisions for domain-adaptive video reasoning.
    }
    \label{fig:main1}\vspace{-0.1in}
\end{figure*}

\section{Introduction}
\label{sec:intro}

Understanding complex video content requires integrating rich spatiotemporal cues and adapting to diverse domain-specific reasoning needs from cinematic narratives, egocentric recordings, to indoor scenes~\cite{fusier2007video,huang2018makes,buch2022revisiting,lin2023univtg,chen2024sharegpt4video,li2024mvbench}.
Models should acquire and integrate a wide range of distinct reasoning skills, such as temporal grounding, spatial relationship recognition, and multi-step planning.

Recent work has extended chain-of-thought (CoT) reasoning~\cite{wei2023chainofthoughtpromptingelicitsreasoning,kojima2022large} to multimodal large language models (MLLMs) for video understanding~\cite{fei2024video,feng2025video,li2025videochat,liu2025videomind,zhi2025videoagent2}.
However, most prior approaches rely on fixed, general-purpose reasoning traces that are insensitive to domain-specific skills.
\cref{fig:main1} (left)
shows a t-SNE~\cite{vanDerMaaten2008tsne} plot of embeddings of questions from different video datasets, where questions from the same datasets are strongly clustered as they require shared skills/domains.
For example, models pretrained on general corpora such as 
LLaVA-Video-178K~\cite{zhang2024videoinstructiontuningsynthetic}
often lack the nuanced narrative understanding needed in CinePile~\cite{rawal2024cinepile}.
This limits their ability to generalize to unseen domains or specialized skills.

To address this, we propose \textsc{Video-Skill-CoT} (aka \textsc{Video-SkoT}), a novel video understanding framework for creating and leveraging skill-aware CoT supervision, helping effective domain adaptation of MLLMs (\cref{sec:method}).
As shown in~\cref{fig:main1} (Right), \ours{} consists of two main components.
First, in skill-based CoT annotation (\cref{sec:data_collection}), we introduce a method to automatically construct high-quality, skill-conditioned CoT rationales for video QA tasks. Given a training question, we first extract high-level reasoning skill descriptions (e.g., ``Determine object location relative to a person's orientation'' and ``Inferring emotional state from expressions and body language''), then cluster them into a shared skill taxonomy (\cref{fig:main1} Right-(a)).
Then, each question is annotated with its top-K relevant skills and used to generate a multi-step CoT annotation conditioned on these skills (\cref{fig:main1} Right-(b)).
This enables diverse and domain-relevant reasoning traces without requiring manual annotation. 

Once we have prepared the skill-based CoT annotations,
in skill-specific expert learning (\cref{sec:multi-lora} and \cref{fig:main1} Right-(c)),
we train skill-specialized expert models with multiple LoRAs~\cite{hu2022lora}.
Each expert specializes in a specific set of reasoning capabilities, determined by a predefined group of related questions. During inference, the model routes each input to the expert aligned with the most relevant question group.

We evaluate \ours{} on three video QA datasets with diverse domains (E.T.-Bench~\cite{liu2024etbench}, VSI-Bench~\cite{yang2024vsi-bench}, and CinePile~\cite{rawal2024cinepile}),
where \ours{} consistently improves over strong baselines, showcasing its strong domain adaptation capabilities.
We also present ablation studies on our design choices and visualize the learned domain-specific skills to validate the effectiveness and interpretability of our skill-guided reasoning framework.

\section{Related Work}
\label{sec:related_work}

\paragraph{Video Understanding with MLLMs.}
Prior video understanding models focused on pretraining strategies~\cite{sun2019videobertjointmodelvideo,li2020herohierarchicalencodervideolanguage,lei2021less}.
Recent work incorporates CoT reasoning~\cite{kojima2022large,wei2023chainofthoughtpromptingelicitsreasoning} from the NLP domain and studies how to collect and learn to generate such CoT reasoning for different video understanding tasks~\cite{fei2024video,li2025videochat,liu2025videomind,zhi2025videoagent2}.
Unlike these methods, which often struggle with comprehending videos without explicit skill-specific guidance, our approach introduces a skill-aware reasoning framework incorporating question-adaptive skill selection and skill-guided CoT supervision.

\paragraph{Skill-specific Expert Learning.}
Modular and expert-based architectures have been widely explored to improve parameter efficiency and mitigate interference in multi-task and multi-domain settings,
where each expert learns different knowledge. 
Mixture-of-experts (MoE) frameworks dynamically route inputs to expert sub-networks~\cite{Shazeer2017Mixture},
while adapter-based methods introduce lightweight, task-specific modules into pretrained models~\cite{houlsby2019parameterefficienttransferlearningnlp,hu2022lora}.
\citet{li2024selma} studies learning skill-specific expert diffusion models for the text-to-image generation task.
A concurrent work, \citet{liu2025videomind} studies a multi-agent system where each agent is implemented as a LoRA~\cite{hu2022lora} expert.
While \citet{liu2025videomind} relies on predefined expert roles (planner, grounder, verifier, and answerer), specific architectures, and manually curated role-specific annotations, our expert framework flexibly adapts to any video understanding dataset by automatically discovering and leveraging relevant reasoning skills.

\section{\textsc{Video-Skill-CoT}}
\label{sec:method}

\subsection{Problem Setup}

Given a video $v$ and a question $q$, our objective is to produce both an answer ${a} $ and a reasoning trace $r$ that offers an interpretable, step-by-step justification. Prior work typically uses a single MLLM $f$ to generate these:
$\{r;\,a\}{=}f(q, v)$.

In contrast, \ours{} decomposes the reasoning process into two stages:
First, given $q$, we select the most relevant expert $e\in\{1,\dots, N^{\mathrm{experts}}\}$ based on the set of pre-defined question groups and predicted required skills.
Next, a \textit{skill-specific expert MLLM $f^{e}$} then generates a \textit{skill-guided reasoning trace $r^{s}$} along with the final answer:
$\{r^{s};\, a\}{=}f^{e}(q, v)$.
We illustrate \ours{} in \cref{fig:main1} (right).

This design enables targeted expert learning and adaptation to diverse reasoning skills in a new video domain.
In the following, we describe
how we automatically construct the skill-based CoT (\cref{sec:data_collection})
and how to train MLLMs with the collected skill-based CoT annotations (\cref{sec:multi-lora}).

\subsection{Skill-based CoT Annotation}
\label{sec:data_collection}
We first construct skill-based CoT rationale annotations for any Video QA dataset, leveraging skill-aware reasoning to enable domain-adaptive video understanding. We perform the following two steps for each $(q,v)$ in the training set to obtain skill-conditioned reasoning traces.

\paragraph{Step 1: Skill Description \& Clustering (\cref{fig:main1} Right-(a)).}
We define a \textit{skill} as a shared, high-level reasoning capability (e.g., temporal ordering, visual counting, spatial understanding) that recurs across multiple video QA examples within a specific domain.
For each question $q$, we prompt an MLLM to describe what kind of skill is necessary to answer it (e.g., \textit{“Estimate distance between two objects using visual cues”}).
Then, we encode all skill descriptions into text embeddings and perform $k$-means clustering (with $k{=}N^{\text{skills}}{=}10$) to form a shared skill taxonomy. Each cluster centroid represents a prototypical skill.

\paragraph{Step 2: Skill-based CoT Collection (\cref{fig:main1} Right-(b)).}
For each $(q, v)$ pair, we generate a multi-step reasoning trace conditioned on the descriptions of the top 3
assigned skills, a process we refer to as \textit{Skill Selection}.
Next, we generate the skill-aware CoT $r^s$; We prompt an MLLM to produce intermediate sub-questions and corresponding answers, guided by selected skills from the previous stage. These sub-QA pairs are then merged into a coherent CoT paragraph that explicitly reflects the assigned reasoning skills.
To ensure the quality of the skill-based CoT rationales, we further verify and filter out reasoning steps that are irrelevant to the correct answer using an LLM evaluator.

After these steps, each training example is now annotated with relevant expert labels $e$ and a verified, skill-grounded CoT trace $r^s$. These annotations form the basis for downstream training of skill-specific expert models.

\subsection{Skill-specific Expert Learning}
\label{sec:multi-lora}

As illustrated in \cref{fig:main1} Right-(c),
we perform modularized fine-tuning to learn task-specific knowledge for skill-based CoT training.
Specifically, we first project all questions in training set $D^\text{train}$ into the text embedding space and perform $k$-means clustering (with $k{=}N^{\text{experts}}{=}5$).
Unlike step 2 of \cref{sec:data_collection} where $N^{\text{skills}}$ clusters represent the groups of \textit{skill descriptions}, these $N^{\text{experts}}$ cluster centroids represent the groups of \textit{questions}.
After assigning each training example to its closest $N^{\text{experts}}$, we conduct parameter-efficient training using the corresponding $N^{\text{experts}}$ expert LoRA~\cite{hu2022lora} modules, ensuring task-specific adaptation while minimizing interference across skills.
During test time, we assign each test question by finding the closest question group by finding the closest question embedding centroids.

\paragraph{Training Objective.}

Following previous work \cite{vpd,aotd}, we train an MLLM by minimizing cross-entropy losses for predicting both the answer ($\mathcal{L}_{\text{answer}}$) and CoT tokens ($\mathcal{L}_{\text{CoT}}$), respectively:
\begin{equation}
\begin{split}
\mathcal{L}&=\mathcal{L}_{\text{answer}}+\lambda\mathcal{L}_{\text{CoT}} \\
&=\ell(f(q,v),a)+\lambda\ell(f(q,v),r^s),
\end{split}
\end{equation}
where we find $\lambda$ = 0.5 balances the two losses well.

\begin{figure*}[t]
    \centering
    {
    \includegraphics[width=\textwidth]{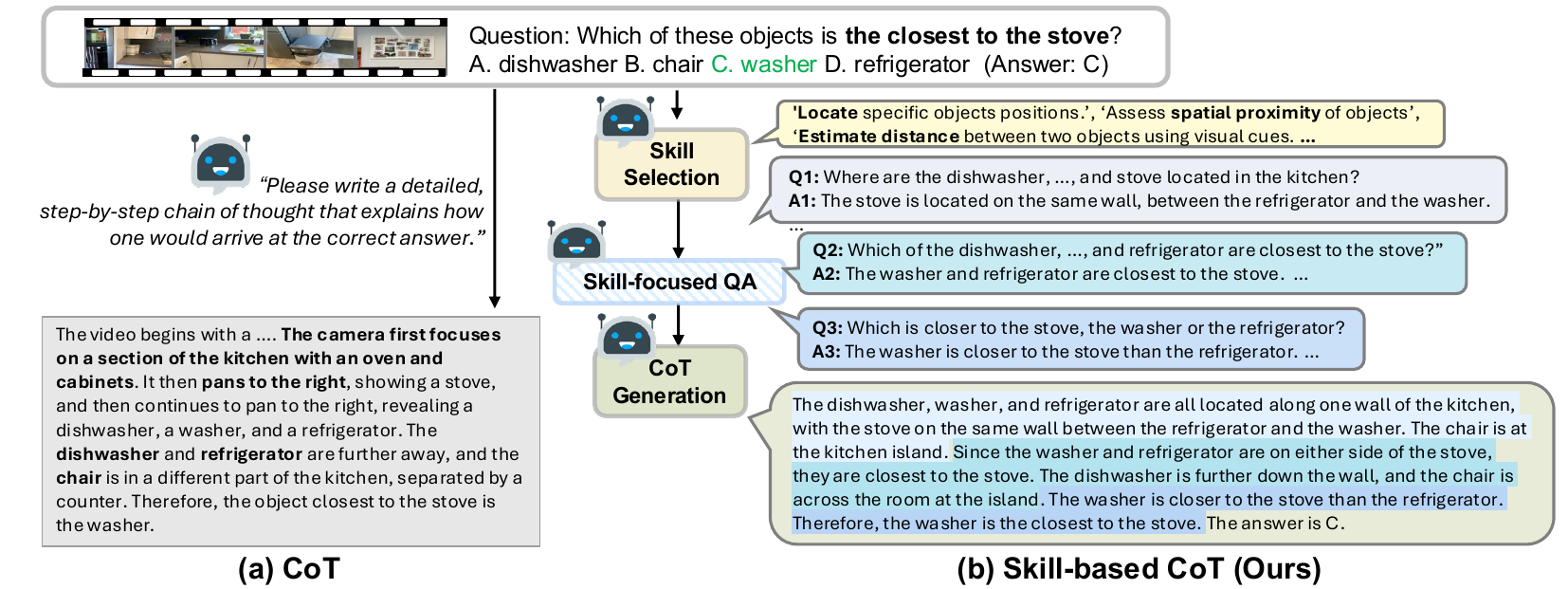}}
    \vspace{-0.1in}
    \caption{
    \textbf{Comparison of CoT annotations: (a) regular CoT and (b) our skill-based CoT.} Additional examples are provided in Appendix~\cref{sec:add_qual_results}.
    }
    \label{fig:data_annotation_examples}
\end{figure*}

\section{Experiments}
\label{sec:experiments}
\subsection{Experiment Setups}
\paragraph{Implementation Details.}
To obtain text embeddings (of skill taxonomy in \cref{sec:data_collection} and of questions in \cref{sec:multi-lora}), we use
\texttt{all-mpnet-base-v2} SentenceTransformers~\cite{reimers-2019-sentence-bert} implementation.
We use LLaVa-Video (7B)~\cite{zhang2024videoinstructiontuningsynthetic} as a main backbone model.
Additional training details including hyperparameters, the specific MLLMs and LLMs used at each stage, as well as results with the Qwen2.5-VL (7B) backbone are provided in Appendix~\cref{sec:detail_cot_generation,sec:detail_skill_description,sec:qwen}.

\paragraph{Datasets and Baselines.}

We experiment with three different video understanding benchmarks with distinct domains:
E.T.Bench~\cite{liu2024etbench} (temporal understanding),
VSI-Bench~\cite{yang2024vsi-bench} (spatial understanding),
and
CinePile~\cite{rawal2024cinepile} (movie narrative understanding).
For multiple-choice questions, we report the average accuracy. For temporal captioning tasks in E.T.Bench, we use the benchmark's official evaluation script. 
Baseline MLLMs include mPLUG-Owl~\cite{ye2024mplug},
Video-ChatGPT~\cite{maaz2023video},
Video-LLaMA2~\cite{zhang2023video}, LLaVa-OneVision~\cite{li2024llava}, and LLaVa-Video~\cite{zhang2024videoinstructiontuningsynthetic},
GPT4o~\cite{gpt4o} and Gemini 1.5 Flash, Pro~\cite{team2024gemini}.
Additional details are provided in the Appendix \cref{sec:detail_training}.

\subsection{Quantitative Evaluation}

\paragraph{Comparison to Baselines.}

\begin{table}[t]
\centering
\small
\renewcommand{\arraystretch}{1.3}
\vspace{-0.05in}\resizebox{0.99\linewidth}{!}{%
\begin{tabular}{lc |ccc}
\toprule
\textbf{} & \multirow{2}{*}{\textbf{Fine-tuned}} & \textbf{E.T.Bench} & \textbf{VSI} & \textbf{CinePile}\\
& & (Temporal) & (Spatial) & (Movie) \\
\midrule
\textit{\textbf{Closed-source MLLMs}} & & & \\
GPT4o~\cite{gpt4o} & & 24.69 & 34.00 & 56.06 \\
Gemini 1.5 Pro~\cite{team2024gemini} & & 26.73 & 45.40 & 60.12 \\
Gemini 1.5 Flash~\cite{team2024gemini} & & 27.92 & 42.10 &58.75 \\
\midrule 
\textit{\textbf{Open-source MLLMs (7B)}} & & & \\
mPLUG-Owl~\cite{ye2024mplug}  & \xmark    & 11.87 & - & 13.93 \\
Video-ChatGPT~\cite{maaz2023video} &  \xmark & 13.02 & - & 15.08 \\
Video-LLaMA2~\cite{zhang2023video} & \xmark    & 8.30 & - & 44.57 \\
LLaVA-Video~\cite{zhang2024videoinstructiontuningsynthetic}  &  \xmark   & 19.35 & 35.60 & 55.83 \\
LLaVA-Video~\cite{zhang2024videoinstructiontuningsynthetic}   &   \checkmark    & 20.32 & 47.45 & 56.29 \\
\rowcolor{lightblue}
\textbf{Ours}    & \checkmark   &    \textbf{22.21}   & \textbf{53.15}  & \textbf{57.88}\\
\bottomrule
\end{tabular}
}
\caption{
\textbf{Evaluation results on domain-specific video reasoning benchmarks.}
}
\label{tab:domain_specific_video}
\vspace{-0.1in}
\end{table}

We compare \ours{} to recent MLLM baselines on three video understanding benchmarks (E.T.Bench, VSI-Bench, CinePile) with domains and required skills.
\Cref{tab:domain_specific_video} shows that \ours{} consistently outperforms all baselines, achieving improvements of $+4.10$, $+5.70$, and $+1.59$ over the fine-tuned version of LLaVA-Video on E.T.Bench, VSI-Bench, and CinePile, respectively. These results highlight the effectiveness of our modular, expert-driven framework in enabling domain-adaptive CoT video reasoning by leveraging relevant skills.

\paragraph{Ablation Studies.}

We compare the impact of two key components in our framework: (1) skill-based CoT reasoning and (2) skill-specific expert modules. 
As shown in~\cref{tab:ablation}, our full model, combining both components (Top row), achieves the highest performance. Removing either the skill-specific expert modules (2nd row), the skill-based CoT (3rd row), or both components (last row) consistently leads to performance degradation, highlighting their complementary roles: skill-CoT enables structured reasoning, while expert modules bring modular specialization. This synergy proves essential for improving video understanding.

\begin{table}[t]
\centering
\small 
\renewcommand{\arraystretch}{1.2}
\vspace{-0.05in}
\resizebox{\columnwidth}{!}{%
\begin{tabular}{c c |c}
\toprule
\textbf{Skill-CoT} (\cref{sec:data_collection}) &  \textbf{Skill-specific Experts} (\cref{sec:multi-lora})  & \textbf{3-Task Avg.}\\
\midrule
\rowcolor{lightblue}
\ding{52} & \ding{52} & \textbf{44.41} \\

\ding{52} & -         & 42.91 \\

-         & \ding{52} & 38.53 \\

-         & -         & 41.04 \\

\bottomrule
\end{tabular}
}
\caption{\textbf{Ablation studies on removing the main components}: Skill-CoT and skill-specific experts.}
\label{tab:ablation}
\end{table}

\begin{table}[t]
\centering
\resizebox{\columnwidth}{!}{%
\begin{tabular}{lccc}
\toprule
\textbf{Criterion} & \textbf{Regular CoT} & \textbf{Skill CoT (Ours)} & \textbf{$\Delta$ (Ours - Regular)} \\
\midrule
Correctness & 2.88 $\pm$ 1.61 & 4.97 $\pm$ 0.16 & +2.09 \\
Relevance   & 3.16 $\pm$ 1.55 & 4.80 $\pm$ 0.59 & +1.64 \\
Coherence   & 2.80 $\pm$ 1.42 & 4.73 $\pm$ 0.64 & +1.93 \\
\bottomrule
\end{tabular}
}
\caption{
\textbf{Human evaluation results comparing Regular CoT and Skill-CoT.} Scores are reported as mean $\pm$ standard deviation.}
\vspace{-0.5cm}
\label{tab:human}
\end{table}

\begin{figure*}[t]
\vspace{-0.5cm}
    \centering
    {
    \includegraphics[width=\textwidth]{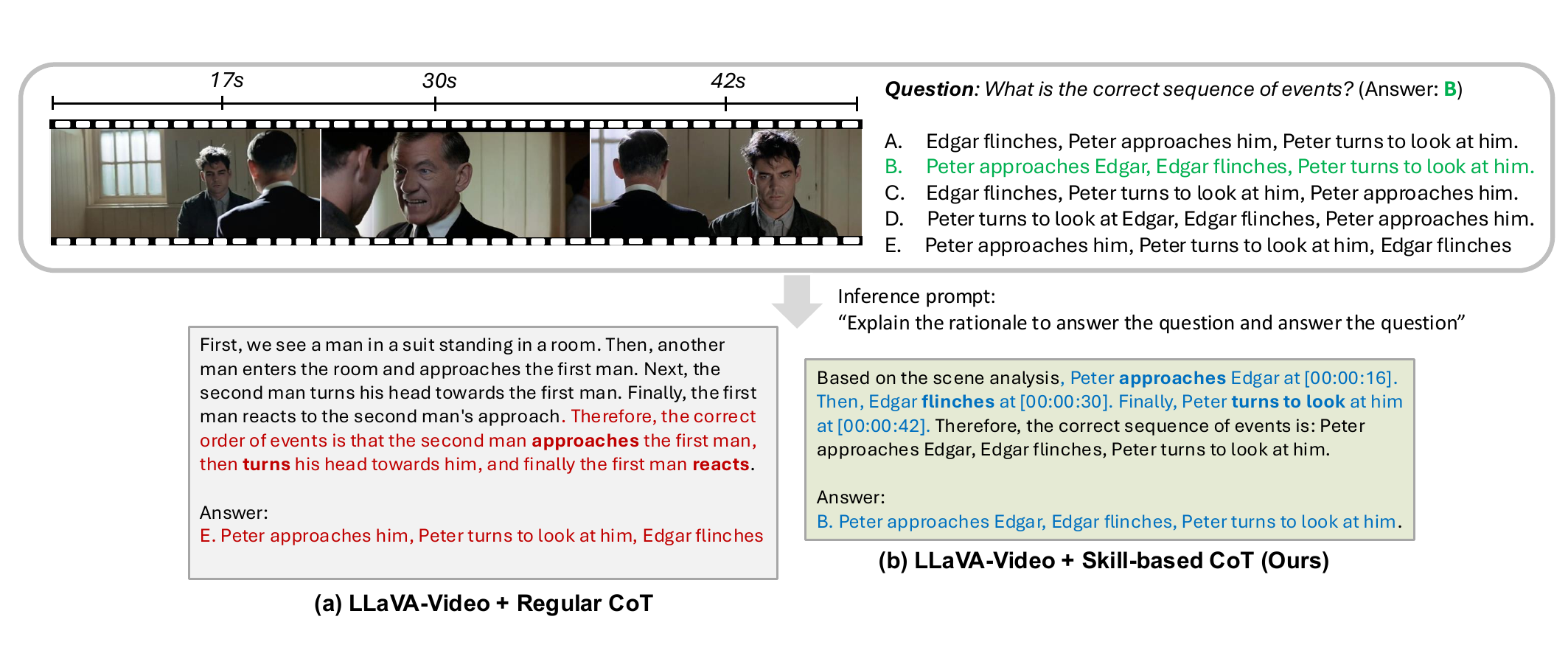}}
    \vspace{-0.3in}
    \caption{
    \textbf{
    Inference output comparison: (a) LLaVA-Video trained with regular CoT and (b) LLaVA-Video trained with our skill-based CoT.} \ours{} successfully generates temporally grounded and precise rationales that more effectively support accurate answer generation.
    }
    \label{fig:qual_inference}
\end{figure*}

\paragraph{Human Evaluation.}
We conduct a human evaluation with five researchers who are familiar with the relevant field, where 15 randomly selected questions were assessed by comparing regular CoT and the proposed Skill-based CoT. 
Each explanation is rated on a 1–5 Likert scale (5 = best, 1 = worst) across three dimensions: Correctness (factual accuracy), Relevance (task appropriateness), and Coherence (clarity and logical flow). 
As shown in \cref{tab:human}, Skill-based CoT consistently outperforms regular CoT across all criteria, with substantial gains in correctness, relevance, and coherence, confirming that our method produces explanations that are more accurate, aligned, and easier to follow.
These results provide strong evidence that Skill-based CoT produces explanations that are not only more accurate but also more relevant and human-readable.

\subsection{Qualitative Analysis}

\paragraph{Regular CoT vs. Skill-based CoT.}
\cref{fig:data_annotation_examples} compares the different annotated CoTs from the regular CoT and our skill-based CoT. Given a question about which object is closest to the stove, the regular CoT (left) offers a linear, scene-based narration that lacks structure and includes irrelevant details (``Camera first focuses ... it then pans to the right ...''), making it often harder to extract key spatial information.
In contrast, our skill-based CoT starts by identifying relevant skills (e.g., spatial proximity) and breaking the task into focused sub-questions, like comparing the washer and refrigerator.

\paragraph{Inference rationale comparison}
We compare the inference-time rationales generated by LLaVA-Video trained with (a) regular CoT and (b) the proposed skill-based CoT. During inference, we prompt each model with: \textit{“Explain the rationale to answer the question and answer the question.”} As shown in \cref{fig:qual_inference}, the model trained with regular CoT produces an incorrect reasoning process, ultimately leading to a wrong answer. In contrast, \ours{} successfully generates temporally grounded and precise rationales that more effectively support accurate answer generation.

\section{Conclusion}
\label{sec:conclusion}
We propose \ours{}, a novel video understanding framework for effective domain adaptation of MLLMs.
We propose to automatically collect skill-specific CoT annotations from video QA datasets and construct a skill-based reasoning pipeline that combines a lightweight skill assigner with a collection of LoRA-based expert adapters.
Empirical results on three diverse benchmarks demonstrate consistent gains of \ours{} over strong baselines, highlighting the enhanced quality of our reasoning traces.

\section*{Limitations}
Our proposed framework demonstrates strong video reasoning capabilities, generating fine-grained, domain-adaptive rationales based on required skills.
However, it may still produce occasional inaccuracies or hallucinations~\cite{liu2023mitigating, wang2024mitigating, zhou2024analyzing} in its text outputs.
Additionally, the overall performance is influenced by the underlying pre-trained backbones, namely, the LLM~\cite{achiam2023gpt4} and MLLM~\cite{team2024gemini} used.
Nonetheless, we highlight that \ours{} can benefit further from future advancements in LLM and MLLM backbones.

\section*{Acknowledgments}
This work was supported by DARPA ECOLE Program No. HR00112390060, NSF-AI Engage Institute DRL-2112635, DARPA Machine Commonsense (MCS) Grant N6600119-2-4031, ARO Award W911NF2110220, ONR Grant N00014-23-1-2356, a Bloomberg Data Science PhD Fellowship,
and the Accelerate Foundation Models Research program. The views contained in this article are those of the authors and not of the funding agency.

{
\bibliography{ref}
}

\newpage

\appendix

\section*{Appendix}

\addcontentsline{toc}{section}{Appendix Table of Contents}
\startcontents[appendix]
\printcontents[appendix]{l}{1}{\setcounter{tocdepth}{2}}

\section{\ours{} Implementation Details}

\subsection{Details of skill description \& clustering}
\label{sec:detail_skill_description}

\textbf{Skill Description.} To extract skill descriptions given the training dataset, we prompt GPT-4\footnote{gpt-4-32k} with its questions and answers. (The prompt is provided in \cref{fig:prompt_skill_abstraction})
Each extracted skill is written as a concise skill phrase (6–12 words), preserving the core visual or temporal reasoning concept. 
Here, we intentionally exclude audio-based cues (e.g., sound, speech, or music) in this process.
Specific object names (e.g., "TV", "sofa", "John") are replaced with generic terms, and vague terms (e.g., "reasoning", "analysis") are avoided to enhance clarity. 
We also provide the exact name of the skills in \cref{tab:skill_list}.

\subsection{Details of skill-based CoT generation}
\label{sec:detail_cot_generation}
For skill-based CoT generation, we utilize Gemini-2.0 Flash with video input. As illustrated in \cref{fig:prompt_gemini}, we first prompt Gemini-2.0 to identify the relevant skills and generate corresponding sub-questions and answers. Then, we construct step-by-step reasoning based on this output. Finally, we use GPT-4 to filter and verify the reasoning by assessing its relevance to the ground-truth answers using \cref{fig:prompt_filtering} as a prompt.

\begin{table*}[ht]
\centering 
\small 
\renewcommand{\arraystretch}{1.3}
\resizebox{.7\textwidth}{!}{%
\begin{tabular}{c|l}
\hline
 & \multicolumn{1}{c}{Skill descriptions} \\ \hline
VSI-Bench & \begin{tabular}[c]{@{}l@{}}"Determine object location relative to a person's orientation.",\\     "Locate specific objects and identify their positions in the scene.",\\     "Identify the closest object to a reference point.",\\     "Recognize objects based on shape, color, and features.",\\     "Identify and count distinct objects based on location and appearance.",\\     "Assess spatial proximity of objects relative to a reference point.",\\     "Identify initial appearances of objects in the video timeline.",\\     "Estimate distance between two objects using visual cues.",\\     "Determine room boundaries using structural elements like walls and floors.",\\     "Identify sequential order of object appearances in a scene."\end{tabular} \\ \hline

ET-Bench & \begin{tabular}[c]{@{}l@{}}"Identify head orientation and gaze direction to infer focus.",\\    "Identify spatial proximity between people in the scene.",\\    "Identifying the timestamp of an action in a scene.",\\    "Detect object using shape, texture, and visual features.",\\    "Track individuals interacting with an object and their actions.",\\    "Identify a person performing an action involving an object.",\\    "Identifying the moment an action occurs in a scene.",\\    "Detect a person using body shape, face, and clothing.",\\    "Identify actions through body movements and postures.",\\    "Identify hand movement and physical interaction with an object."\end{tabular} \\  \hline

CinePile & \begin{tabular}[c]{@{}l@{}}"Identify thematic parallels between actions and overarching narrative themes.",\\     "Inferring emotional tone from facial expressions and actions.",\\     "Tracking emotional shifts through expressions and body language changes.",\\     "Identifying interpersonal conflict through observed actions and interactions.",\\     "Inferring symbolic meaning of an object in a scene.",\\     "Inferring emotional state from expressions and body language.",\\     "Track a person's movements and reactions to scene changes.",\\     "Identifying body language among the scene.",\\     "Identify event sequence to infer action context and significance.",\\     "Identifying the main person or subject in the scene."\end{tabular} \\ \hline
\end{tabular}%
}
\caption{\textbf{Detailed skill descriptions from three datasets.} }
\label{tab:skill_list}
\end{table*}

\subsection{Details of training}
\label{sec:detail_training}

\textbf{Training datasets.} 
Instead of using the full video instruction tuning dataset, we randomly sampled 10k and 2.1k examples from ET-Bench and CinePile, respectively. For VSI-Bench, which is intended solely for evaluation and does not provide a training set, we manually split the available data into training and test sets using a 7:3 ratio. We use 3k training dataset for VSI-Bench. 

\noindent\textbf{Hyperparameters.}
For training, we set the learning rate as 1e-5 and the batch size as $1$. For LoRA, we use rank $32$. 
We set 1 epoch for ET-Bench training and 3 epochs for the other two datasets. 
For other parameters, we use the default setup of LLaVA Video. 
We use 4 A6000 GPUs for training.

\subsection{Prompts}
\label{sec:prompt}
\begin{figure*}[t]
    \centering
    {
    \includegraphics[width=\textwidth]{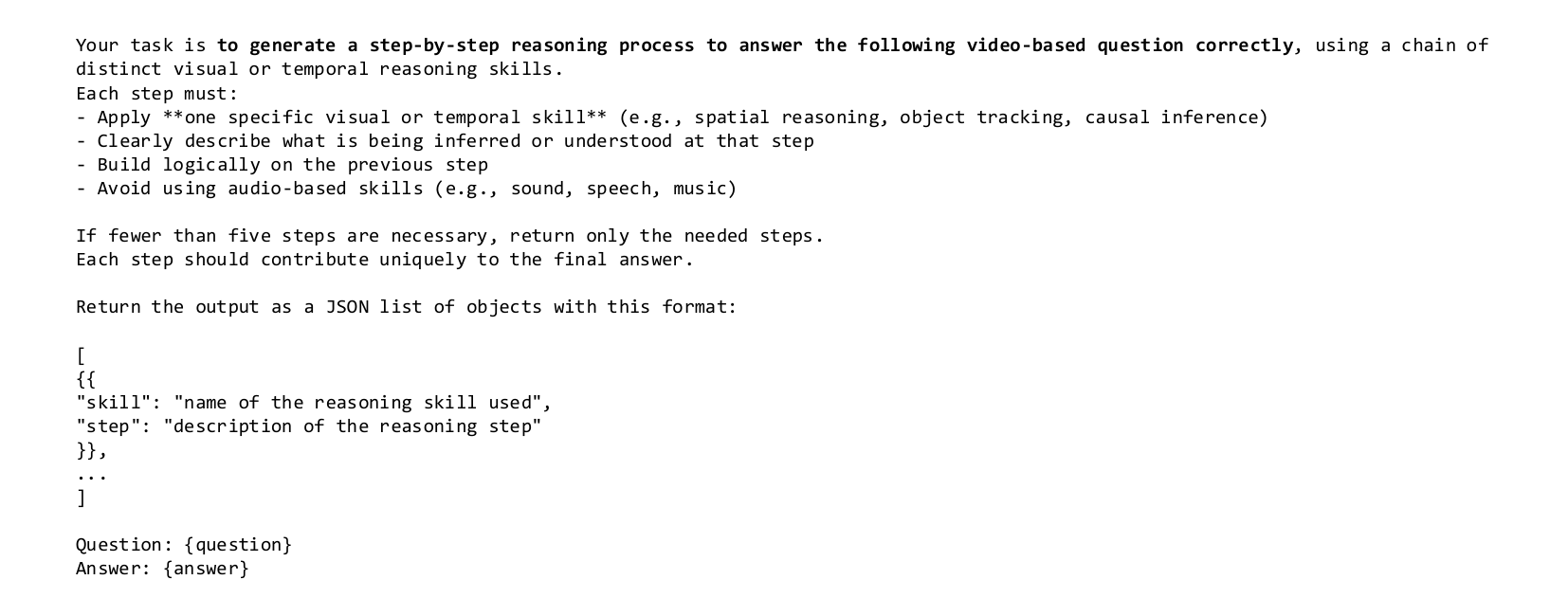}}
    \caption{
    \textbf{\textbf{Prompt for Skill Description}}
    }
    \label{fig:prompt_skill_abstraction}\vspace{-0.1in}
\end{figure*}

\begin{figure*}[t]
    \centering
    {
    \includegraphics[width=\textwidth]{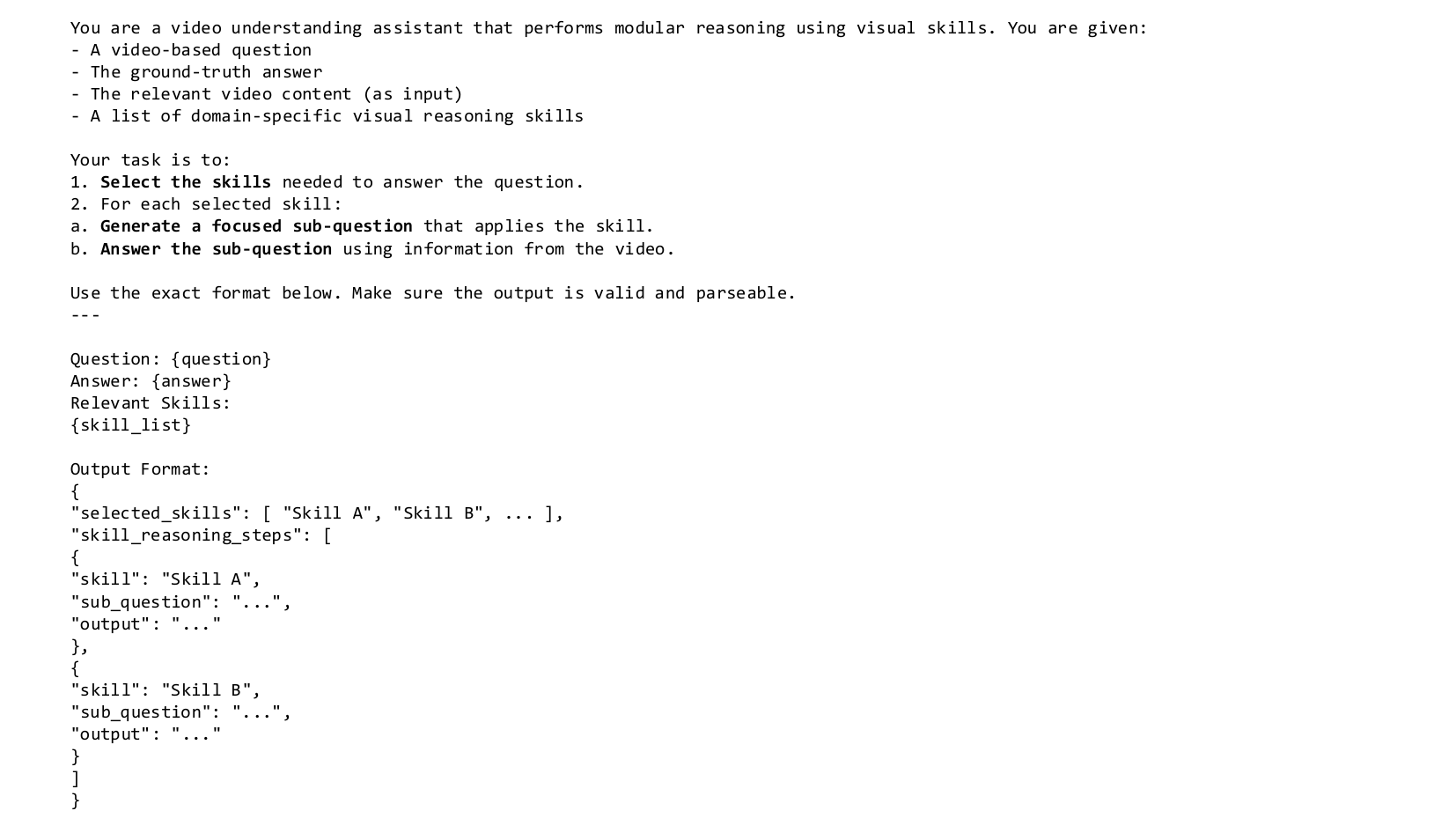}}
    \caption{
    \textbf{\textbf{Prompt for skill selection and sub-QA generation}}
    }
    \label{fig:prompt_gemini}\vspace{-0.1in}
\end{figure*}

\begin{figure*}[t]
    \centering
    {
    \includegraphics[width=\textwidth]{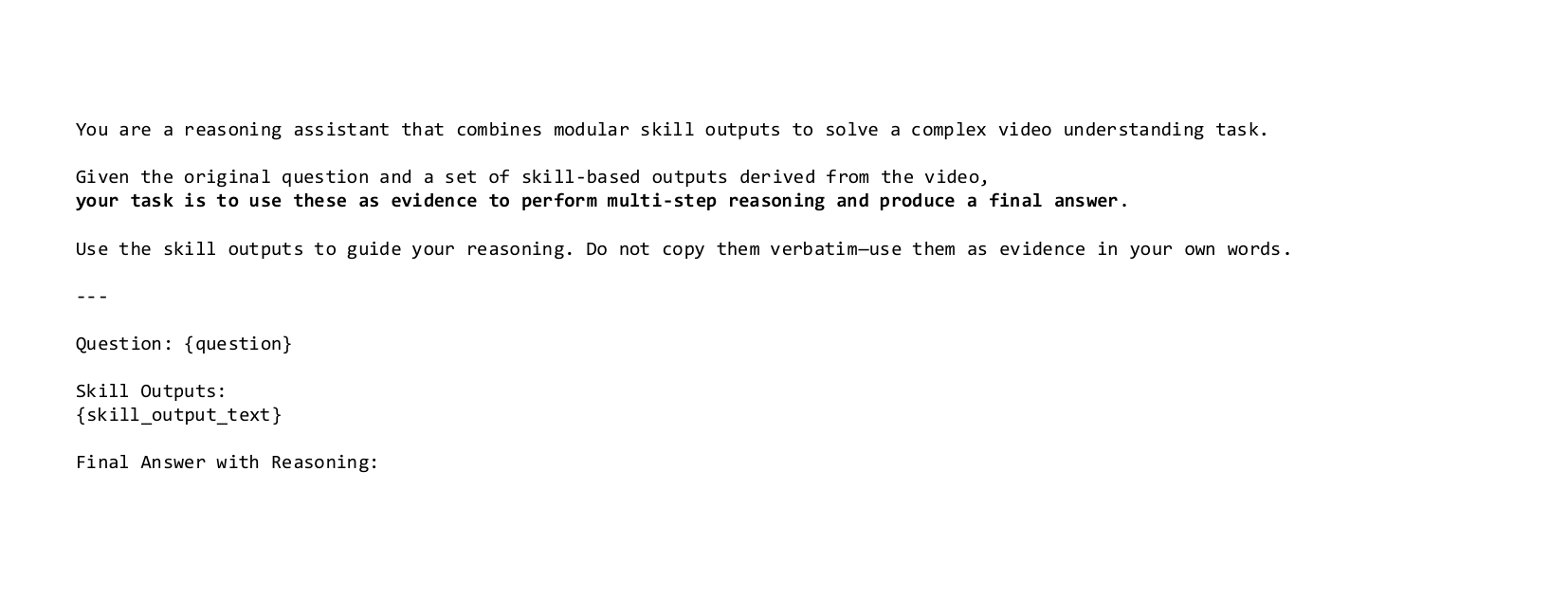}}
    \caption{
    \textbf{\textbf{Prompt for skill-based CoT generation}}
    }
    \label{fig:prompt_merging}\vspace{-0.1in}
\end{figure*}

\begin{figure*}[t]
    \centering
    {
    \includegraphics[width=\textwidth]{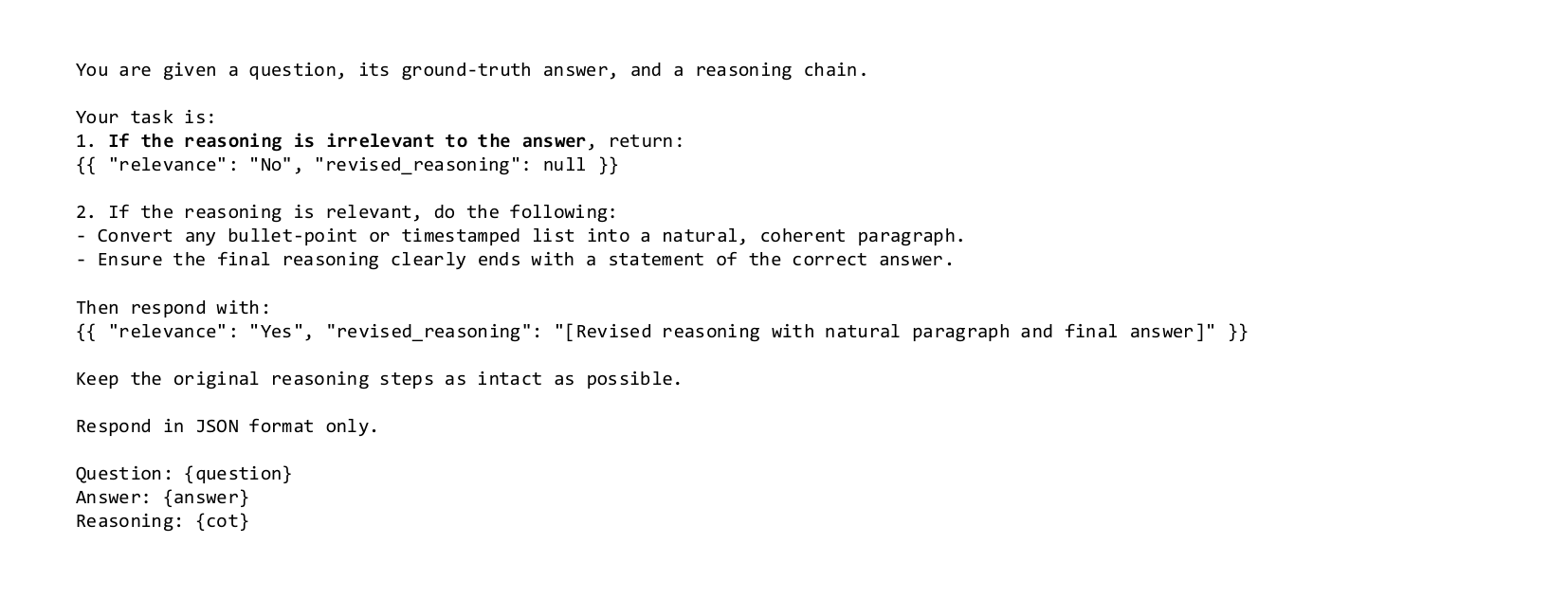}}
    \caption{
    \textbf{\textbf{Prompt for CoT filtering}}
    }
    \label{fig:prompt_filtering}\vspace{-0.1in}
\end{figure*}

\begin{figure*}[t]
    \centering
    {
    \includegraphics[width=\textwidth]{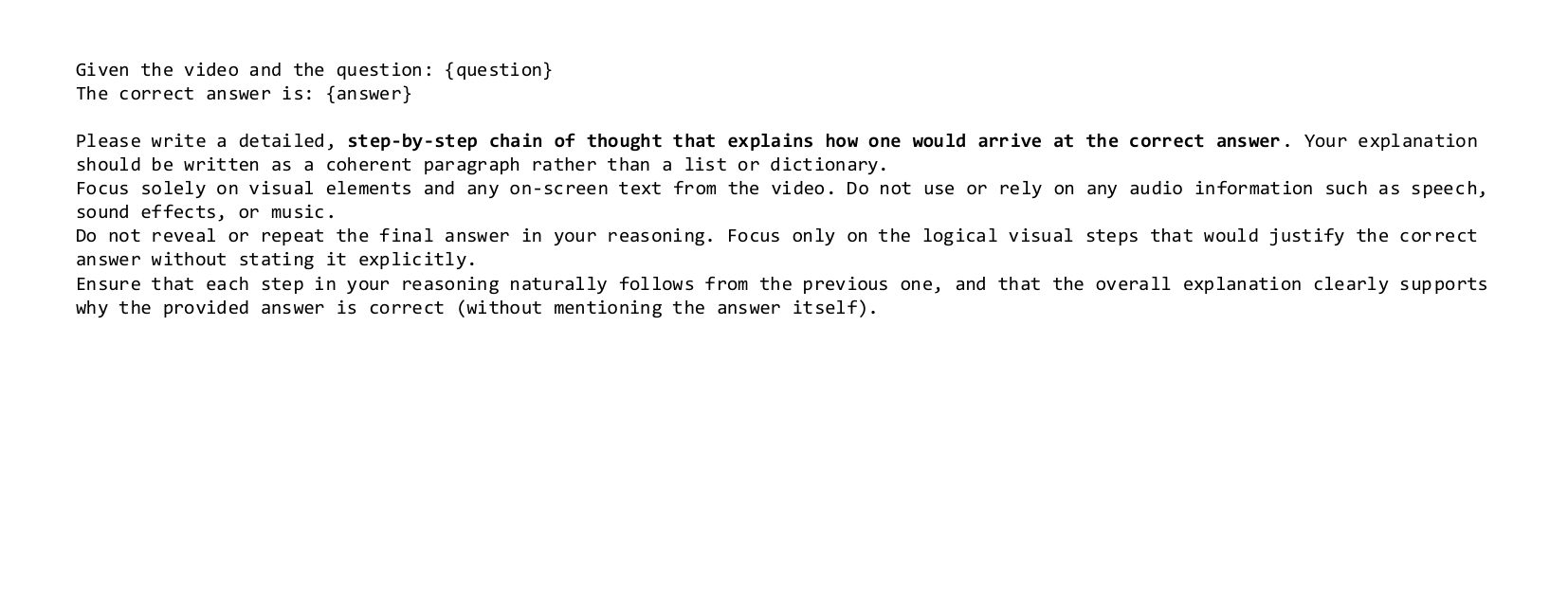}}
    \caption{
    \textbf{\textbf{Prompt for regular CoT generation}}
    }
    \label{fig:prompt_regular_cot_generation}\vspace{-0.1in}
\end{figure*}

In \cref{fig:prompt_skill_abstraction,fig:prompt_gemini,fig:prompt_merging,fig:prompt_filtering}, we attach prompts for skill-based CoT annotation. 
We also attach prompt to generate regular CoT in \cref{fig:prompt_regular_cot_generation}.

\begin{table*}[ht]
\centering 
\small 
\renewcommand{\arraystretch}{1.2}
\resizebox{.95\textwidth}{!}{
\begin{tabular}{l|cccccccc|c}
\toprule
 & Appr. Order & Rel. Dist. & Route Plan & Rel. Dir. & Abs. Dist. & Obj. Count & Obj. Size & Room Size & Avg \\
 \midrule
LLaVA-Video & 31.28 & 41.57 & 35.59 & 46.05 & 10.78 & 52.24 & 47.54 & 19.77 & 35.60 \\
LLaVA-Video (fine-tune) & 54.16 & 36.97 & 41.66 & 42.26 & 29.71 & 56.89 & 70.16 & 54.51 & 47.45 \\
Single-LoRA + Regular-CoT & 68.75 & 36.97 & 41.66 & 43.07 & 31.37 & 62.62 & 70.22 & 65.96 & 52.58 \\
Single-LoRA + Skill-CoT & 64.58 & 42.01 & 36.11 & 44.45 & 33.11 & 56.89 & 74.27 & 64.35 & 52.97 \\
Multi-LoRA + Regular-CoT & 56.25 & 34.45 & 41.66 & 39.53 & 12.24 & 43.00 & 55.11 & 23.87 & 38.27 \\ 
\midrule
\rowcolor{lightblue}
\ours{} (Ours) & 68.75 & 36.13 & 50.00 & 47.69 & 32.13 & 62.61 & 70.83 & 57.09 & 53.15 \\

\bottomrule

\end{tabular}
}
\caption{\textbf{Detailed VSI-Bench Results}.}
\label{tab:detail_vsi}
\end{table*}

\begin{table*}[ht]
\centering 
\small 
\renewcommand{\arraystretch}{1.2}
\resizebox{.95\textwidth}{!}{
\begin{tabular}{l|cccccccccccccc|c}
\toprule
 & RAR & EVC & RVQ & TVG & ERM & TAL & EVS & VHD & DVC (F1) & DVC (Sim) & SLC (F1) & SLC (Sim) & TEM & GVQ & Avg \\
 \midrule
LLaVA-Video & 41.6 & 38.8 & 56.6 & 8.2 & 1.8 & 14.0 & 14.8 & 28.2 & 20.1 & 10.0 & 11.5 & 8.1 & 15.7 & 1.5 & 19.3 \\
LLaVA-Video (fine-tune) & 44.8 & 34.6 & 58.2 & 9.4 & 2.2 & 12.0 & 9.2 & 29.7 & 28.6 & 14.8 & 10.1 & 10.2 & 18.6 & 2.1 & 20.3 \\
Single-LoRA + Regular-CoT & 43.6 & 37.8 & 58.6 & 10.7 & 1.9 & 13.7 & 7.5 & 33.8 & 14 & 12.2 & 6.5 & 10.6 & 12.4 & 3.1 & 19.0 \\
Single-LoRA + Skill-CoT & 43.2 & 40.8 & 56.8 & 10.0 & 1.7 & 15.2 & 7.1 & 30.2 & 17.3 & 16.5 & 10.0 & 10.6 & 11.2 & 1 & 19.4 \\
Multi-LoRA + Regular-CoT & 43.2 & 40.6 & 59.8 & 6.9 & 1.9 & 16.5 & 11.1 & 31.1 & 30.3 & 16.4 & 6.2 & 9.5 & 16.6 & 1.0 & 20.7 \\
\midrule
\rowcolor{lightblue}
\ours{} (Ours) & 49.0 & 41.2 & 59.4 & 15.8 & 2.4 & 16.4 & 8.4 & 35.5 & 27.0 & 15.0 & 11.9 & 10.0 & 16.5 & 2.4 & 22.2 \\
\bottomrule
\end{tabular}
}
\caption{\textbf{Detailed E.T-Bench Results}.}
\label{tab:detail_et}
\end{table*}

\begin{table*}[ht]
\centering 
\small 
\renewcommand{\arraystretch}{1.2}
\resizebox{.65\textwidth}{!}{
\begin{tabular}{l|ccccc|c}
\toprule
 & CRD & NPA & STA & TH & TEMP & Avg \\
 \midrule
LLaVA Video & 56.78 & 58.53 & 60.31 & 60.52 & 43.02 & 55.83 \\
LLaVA Video (fine-tune) & 57.74 & 58.32 & 60.44 & 61.05 & 43.90 & 56.29 \\
Single-LoRA + Regular CoT & 59.18 & 59.61 & 61.81 & 61.05 & 40.91 & 56.11 \\
Single-LoRA + Skill-CoT & 57.88 & 56.8 & 60.77 & 60.52 & 40.84 & 56.36 \\
Multi-LoRA + Regular CoT & 58.17 & 59.17 & 60.18 & 62.63 & 42.44 & 56.52 \\ 
\midrule
\rowcolor{lightblue}
\ours{} (Ours) & 60.00 & 59.61 & 61.44 & 63.15 & 45.20 & 57.88 \\
\bottomrule

\end{tabular}
}
\caption{\textbf{Detailed Cinepile Results.} We ablate the accuracies across the question categories: TEMP
- Temporal, CRD - Character and Relationship Dynamics, NPA - Narrative and Plot Analysis, STA - Setting and
Technical Analysis, TH - Thematic Exploration. 
}
\label{tab:detail_cine}
\end{table*}

\section{Additional Quantitative Results}
\label{sec:add_quant_results}

\subsection{Per-category performance}
\label{sec:per_category}
In \cref{tab:detail_vsi,tab:detail_cine,tab:detail_et}, we additionally report the per-category performance for each dataset. We also include ablation studies comparing regular CoT vs skill-based CoT, and single-LoRA vs multi-LoRA configurations. \ours{}, which combines skill-based CoT with multi-LoRA training, consistently outperforms across all datasets, showing particularly strong gains on reasoning-intensive tasks such as Route Planning in VSI-Bench and temporal understanding tasks in CinePile.

\subsection{Qwen2.5-VL backbone}
\label{sec:qwen}

\begin{table*}[ht]
\centering 
\small 
\renewcommand{\arraystretch}{1.2}
\resizebox{.95\textwidth}{!}{
\begin{tabular}{l|cccccccc|c}
\toprule
\textbf{Model} & \textbf{Appr. Order} & \textbf{Rel. Dist.} & \textbf{Route Plan} & \textbf{Rel. Dir.} & \textbf{Abs. Dist.} & \textbf{Obj. Count} & \textbf{Obj. Size} & \textbf{Room Size} & \textbf{Avg} \\
\midrule
Qwen2.5-VL & 10.41 & 35.29 & 36.11 & 37.80 & 20.43 & 23.98 & 54.22 & 33.06 & 31.41 \\
Qwen2.5-VL (fine-tune) & 18.75 & 34.45 & 36.11 & 38.65 & 21.88 & 31.74 & 62.55 & 47.74 & 36.48 \\
Single-LoRA + Regular-CoT & 20.83 & 35.29 & 36.11 & 39.92 & 21.73 & 34.56 & 63.33 & 48.06 & 37.48 \\
Single-LoRA + Skill-CoT & 27.08 & 35.61 & \textbf{41.66} & 40.12 & 21.57 & 35.33 & 65.77 & \textbf{48.58} & 39.47 \\
\midrule
\rowcolor{lightblue}
Ours (Multi-LoRA + Skill-CoT) & \textbf{31.28} & \textbf{37.89} & 36.50 & \textbf{43.67} & \textbf{28.12} & \textbf{37.48} & \textbf{67.31} & 43.79 & \textbf{40.76} \\
\bottomrule
\end{tabular}
}
\caption{\textbf{Qwen2.5-VL (7B) Results on VSI-Bench}. }
\label{tab:qwen}
\end{table*}

We further evaluate \ours{} on VSI-Bench using the Qwen2.5-VL (7B)~\cite{bai2025qwen2} backbone. As shown in \cref{tab:qwen}, \ours{} achieves the highest overall performance (40.76 avg), consistently surpassing both regular-CoT and single-LoRA baselines. These results highlight the robustness and effectiveness of \ours{} when applied to a different backbone architecture.

\begin{table*}[t]
\centering 
\small 
\renewcommand{\arraystretch}{1.2}
\resizebox{.70\textwidth}{!}{
\begin{tabular}{l|c|ccccc|c}
\toprule
\textbf{Setting} & \textbf{Training Data} & \textbf{CRD} & \textbf{NPA} & \textbf{STA} & \textbf{TH} & \textbf{TEMP} & \textbf{Avg} \\
\midrule
LLaVA-Video & - & 56.78 & 58.53 & 60.31 & 60.52 & 43.02 & 55.83 \\
LLaVA-Video (fine-tune) & CinePile & 57.74 & 58.32 & 60.44 & 61.05 & 43.90 & 56.29 \\
LLaVA-Video (fine-tune) & ET-Bench & 53.34 & 57.02 & 57.83 & 60.53 & 40.55 & 53.85 \\
Single-LoRA + \textbf{Regular-CoT} & ET-Bench & 56.23 & 60.47 & 58.09 & 63.68 & 38.66 & 55.43 \\
\midrule
\rowcolor{lightblue}
Multi-LoRA + \textbf{Skill-CoT (Ours)} & ET-Bench & 56.02 & 61.25 & 58.88 & 63.33 & 41.57 & \textbf{56.21} \\
\bottomrule
\end{tabular}
}
\caption{\textbf{Cross-dataset evaluation on CinePile}. Source: ET-Bench $\rightarrow$ Target: CinePile. Backbone: LLaVA-Video. 
}
\label{tab:cross}
\end{table*}

\subsection{Cross-dataset generalization}
\label{sec:cross_dataset}
We evaluate cross-domain generalization from ET-Bench (source) to CinePile (target). As shown in \cref{tab:cross}, \ours{} achieves the best average performance (56.21) among ET-Bench–trained variants, performing competitively with the CinePile fine-tuned model (56.29) and surpassing the zero-shot baseline (55.83). 
This highlights the effectiveness of skill-guided reasoning for transfer across domains.

\begin{figure*}[t]
    \centering
    {
    \includegraphics[width=\textwidth]{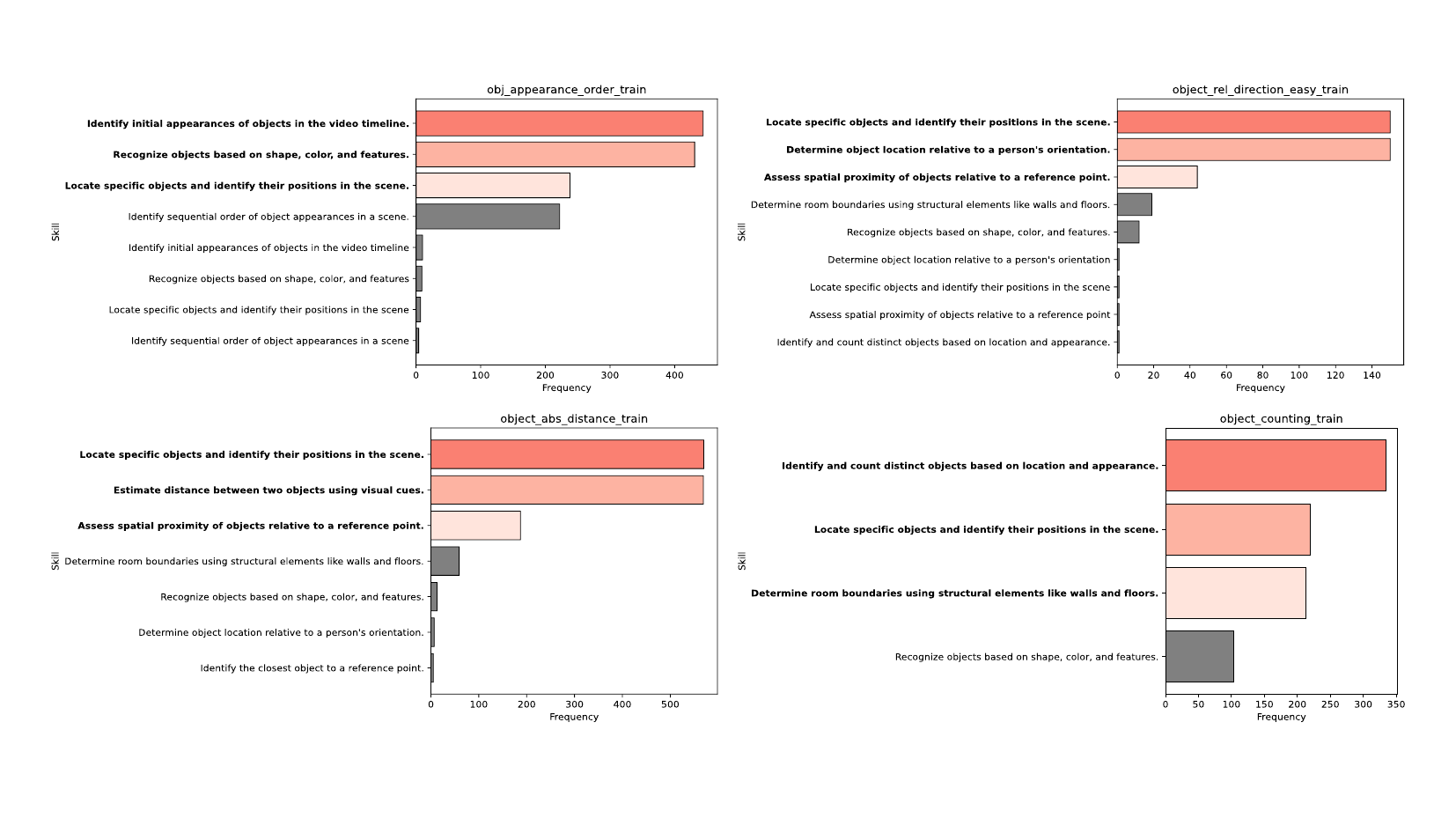}}
    \caption{
    \textbf{Skill selection results of VSI-Bench (1)}
    }
    \label{fig:selected_skills1}
\end{figure*}

\begin{figure*}[t]
    \centering
    {
    \includegraphics[width=\textwidth]{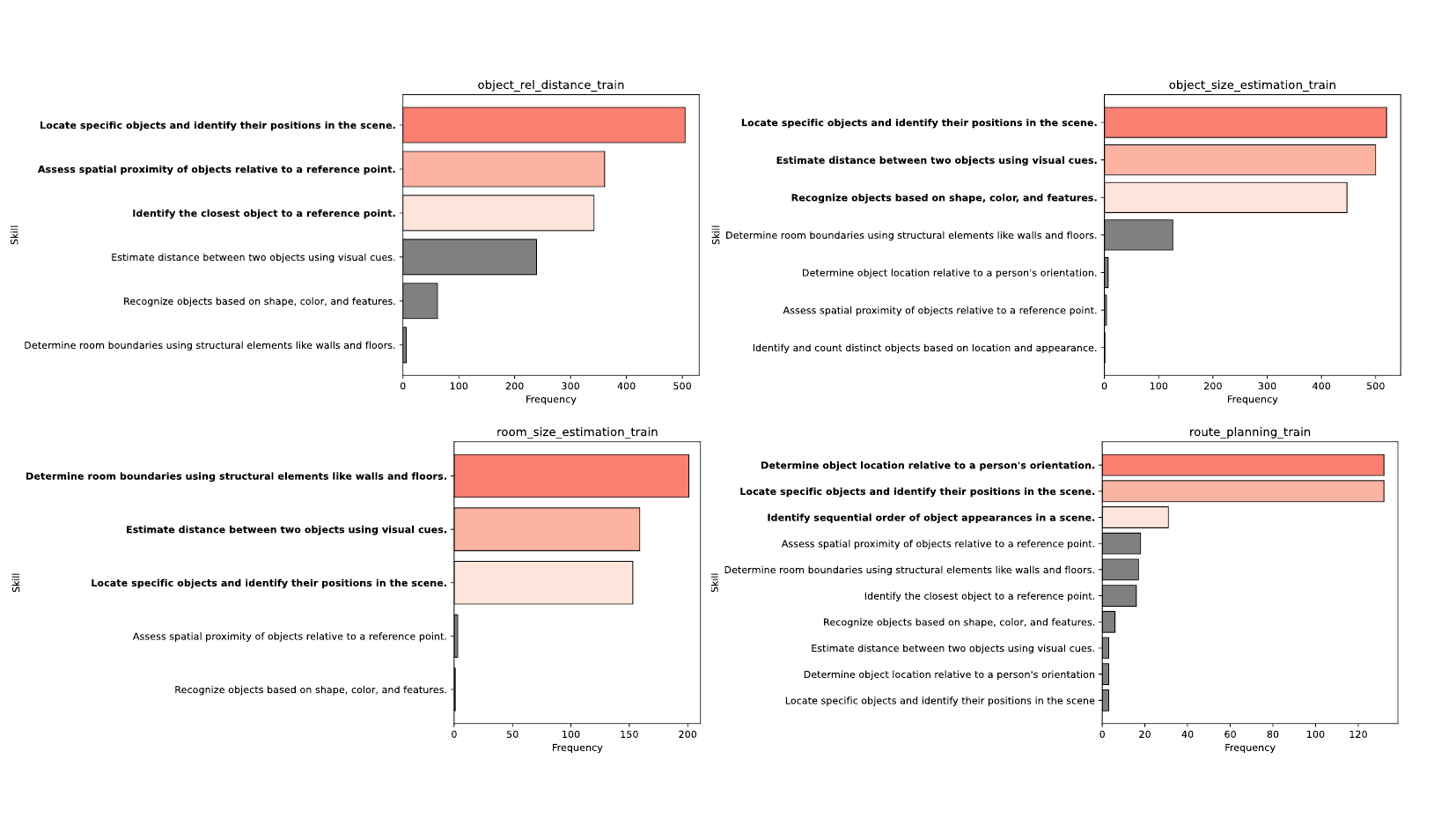}}
    \caption{
    \textbf{\textbf{Skill selection results of VSI-Bench (2)}}
    }
    \label{fig:selected_skills2}\vspace{-0.1in}
\end{figure*}

\section{Additional Qualitative Results}
\label{sec:add_qual_results}

\subsection{Skill descriptions over different datasets}

\begin{figure*}[t]
    \centering
    {
    \includegraphics[width=\textwidth]{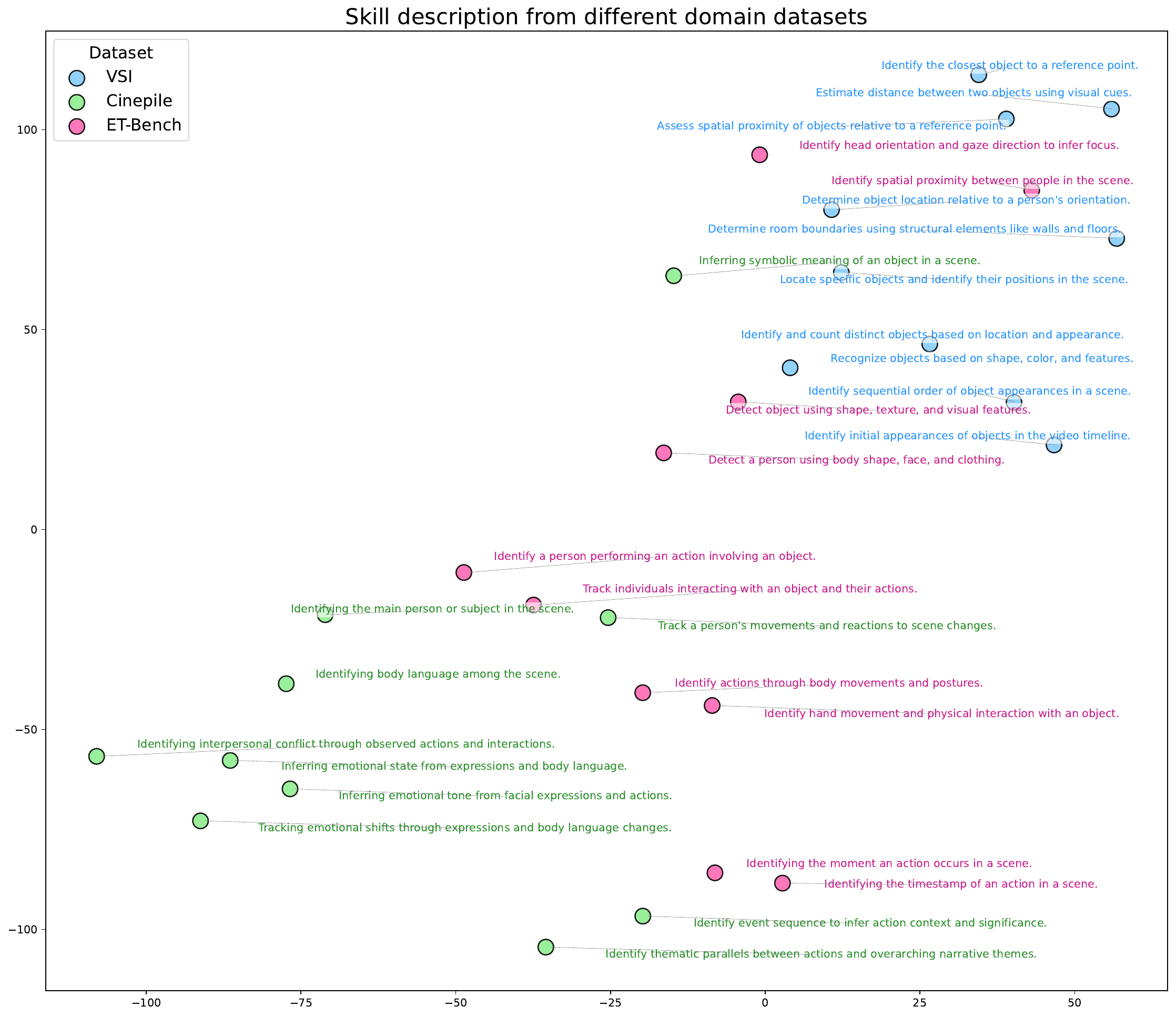}}
    \caption{
    \textbf{Skill description from different domain datasets.} We visualize the skill descriptions for each dataset after performing skill extraction and clustering. (\cref{sec:data_collection})
    }
    \label{fig:skill_description_tsne}\vspace{-0.1in}
\end{figure*}

In \cref{fig:skill_description_tsne}, we visualize the skill descriptions for each dataset after performing skill extraction and clustering (\cref{sec:data_collection}). 
To create the visualization, we first obtain text embeddings using SentenceTransformer and compute $N^{\text{skills}}$ cluster centroids. 
We then apply t-SNE to reduce the dimensionality of the embeddings for visualization purposes. 
The results highlight that each domain-specific dataset emphasizes different skill sets, though certain skills are shared across datasets. For instance, the skill “\textit{Inferring emotional tone from facial expressions and actions}” from CinePile is distinct from “\textit{Estimating distance between two objects in the video timeline}” from VSI-Bench. However, general skills like “\textit{Identifying objects or people}” appear across multiple datasets.
A more detailed list of the extracted skills is provided in \cref{tab:skill_list}. 

\subsection{Selected skills over different video datasets}

In \cref{fig:selected_skills1,fig:selected_skills2}, we present statistics on the selected top 3 assigned skills for each task in VSI-Bench (presented in \cref{sec:data_collection}). As shown in the results, object identification skills are commonly used across tasks. However, each task also requires domain-specific skills. For instance, the Room Size Estimation task necessitates skills such as “\textit{Determining room boundaries using structural elements like walls and floors.}”

\subsection{Additional comparison with regular CoT}
\begin{figure*}[t]
    \centering
    {
    \includegraphics[width=\textwidth]{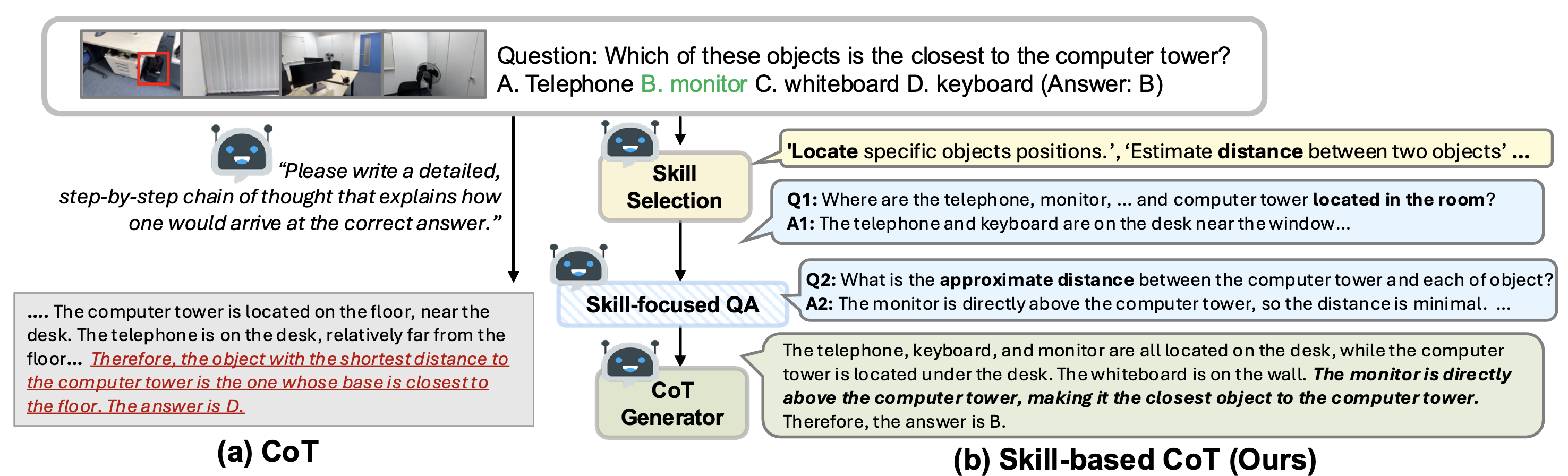}}
    \caption{
    \textbf{\textbf{Additional comparison of CoT annotations: (a) regular CoT and (b) our skill-based CoT.}}
    }
    \label{fig:additional_comparison_regular_cot_generation}\vspace{-0.1in}
\end{figure*}

In \cref{fig:additional_comparison_regular_cot_generation}, we provide additional comparison with regular CoT and ours.

\section{License}
We list the license of the benchmark dataset and models we used. We use these existing artifacts consistently with their intended use. 
\begin{itemize}
    \item LLaVA-Video: \href{https://github.com/LLaVA-VL/LLaVA-NeXT/blob/main/LICENSE}{Apache License 2.0} 
    \item CinePile: \href{https://huggingface.co/datasets/tomg-group-umd/cinepile}{cc-by-nc-sa-4.0}
    \item VSI-Bench: \href{https://huggingface.co/datasets/choosealicense/licenses/blob/main/markdown/apache-2.0.md}{Apache License 2.0}
    \item ET-Bench: \href{https://huggingface.co/datasets/PolyU-ChenLab/ET-Instruct-164K}{cc-by-nc-sa-4.0}
    
\end{itemize}

\end{document}